\documentclass[10pt]{article} 
\usepackage{booktabs}

\usepackage{multirow}
\usepackage{times}
\usepackage{latexsym}
\usepackage{float}
\usepackage{graphicx}
\usepackage{algorithm}
\usepackage{algpseudocode}
\usepackage{amsmath}
\usepackage[T1]{fontenc}
\usepackage[utf8]{inputenc}
\usepackage{microtype}
\usepackage{inconsolata}
\usepackage[table,xcdraw]{xcolor}
\usepackage[preprint]{tmlr}
\usepackage{wrapfig}
\usepackage{bbm}

\usepackage{amsmath,amsfonts,bm}

\def\eqref#1{equation~\ref{#1}}

\def\1{\bm{1}}

\DeclareMathAlphabet{\mathsfit}{\encodingdefault}{\sfdefault}{m}{sl}
\SetMathAlphabet{\mathsfit}{bold}{\encodingdefault}{\sfdefault}{bx}{n}

\usepackage{hyperref}
\usepackage{url}

\def\shownotes{1}  
\ifnum\shownotes=1
\newcommand{\authnote}[2]{[#1: #2]}
\else
\newcommand{\authnote}[2]{}
\fi

\title{Language Models Prefer What They Know: \\Relative Confidence Estimation via Confidence Preferences}

\author{\name Vaishnavi Shrivastava \email vshrivas@cs.stanford.edu \\
      \addr Stanford University
      \AND
      \name Ananya Kumar \email ananya@cs.stanford.edu \\
      \addr Stanford University
      \AND
      \name Percy Liang \email pliang@cs.stanford.edu\\
      \addr Stanford University}

\def\month{MM}  
\def\year{YYYY} 
\def\openreview{\url{https://openreview.net/forum?id=XXXX}} 

\begin{document}

\maketitle

\begin{abstract}
Language models (LMs) should provide reliable confidence estimates to help users detect  mistakes in their outputs and defer to human experts when necessary. Asking a language model to assess its confidence (“Score your confidence from 0-1.”) is a natural way of evaluating its uncertainty. However, models struggle to provide absolute assessments of confidence (i.e. judging confidence in answering a question independent of other questions) and the coarse-grained scores they produce are not useful for evaluating the correctness of their answers. We propose \textit{relative confidence estimation},
where we match up questions against each other and ask the model to make relative judgments of confidence (\textit{``Which question are you more confident in answering correctly?''}). Treating each question as a “player” in a series of matchups against other questions and the model’s preferences as match outcomes, we can use rank aggregation methods like Elo rating and Bradley-Terry to translate the model’s confidence preferences into confidence scores. We evaluate 
relative confidence estimation against absolute confidence estimation and self-consistency confidence methods on five state-of-the-art LMs---GPT-4, GPT-4o, Gemini 1.5 Pro, Claude 3.5 Sonnet, and Llama 3.1 405B---across 14 challenging STEM, social science, and commonsense reasoning question answering tasks.
Our results demonstrate that relative confidence estimation consistently provides more reliable confidence scores than absolute confidence estimation, with average gains of 3.5\% in selective classification AUC over direct absolute confidence estimation methods and 1.7\% over self-consistency approaches across all models and datasets.
\end{abstract}
\section{Introduction}
To ensure users can make informed decisions when interpreting outputs from language models (LMs), it is crucial to develop methods for accurately gauging their confidence. Language models are widely deployed, yet they remain prone to errors in their outputs. For instance, even state-of-the-art models like GPT-4o and Llama 3.1 405B struggle to solve challenging datasets such as GPQA~\citep{Rein2023GPQAAG} and MATH~\citep{math}. To help users detect mistakes in their generations, models should provide reliable confidence estimates, signaling when their responses are more likely to be incorrect. By leveraging these estimates, users can disregard low-confidence answers or seek expert opinions.

Since users primarily engage with chatbots like ChatGPT~\citep{chatgpt} through language, asking language models to gauge their confidence is a natural tool.
A straightforward approach to this is absolute confidence estimation—--asking the model to directly rate its confidence without further context or grounding, e.g., \textit{``How confident are you on a scale of 0-1?''} However,~\cite{surrogate-models} find that absolute confidences can be too coarse-grained and lack discriminative power. For example, GPT-4 produces the same confidence score of 0.9 for 50\% of examples across 12 datasets, limiting its ability to distinguish between correct and incorrect answers. 
\begin{figure}
    \centering
    \includegraphics[width=\linewidth]{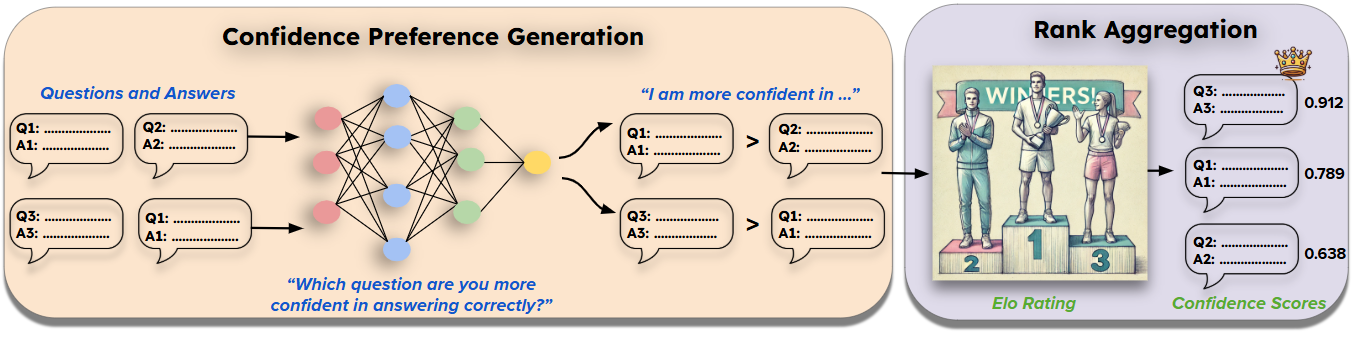} 
    \caption{\textbf{Relative Confidence Estimation.} We first prompt models to elicit their answers to different questions. For each question $q_i$, we match $q_i$ with $n$ other questions $q_j$ and generate confidence preference data. We ask the model to compare its level of confidence in the pair of questions and decide which question it is \textit{more} confident in answering correctly. We treat the questions and answers as ``players'' in these matchups and the confidence preferences as match outcomes. Leveraging rank aggregation techniques used in competitive games, such as Elo rating, we translate the model's confidence preferences into confidence scores.}
    \label{fig:intro-fig}
    \vspace{-0.2in}
\end{figure}
This may be due to a lack of realistic examples of confidence estimation in training data. For example, ~\cite{zhou2023navigating} find that many examples in the Pile dataset use hyperbolic terms like “I am 100\% confident,” rather than providing more nuanced estimates.

We introduce \textit{relative confidence estimation}, as an alternative to absolute confidence estimation. Rather than asking models to rate their confidence on an answer to a single question, we ask them to compare confidence across different questions: \textit{``Which question are you more confident in answering correctly?''}. Relative comparisons are used in many scenarios as an easier alternative to absolute judgments. For instance, in RLHF, annotators assess which generation is better, rather than assigning direct scores~\citep{instruct-gpt}. ~\cite{kadavath2022language} also show that LMs are better at making relative judgments of correctness by comparing multiple sampled outputs, rather than verifying a single generation. To the best of our knowledge, ours is the first study to explore confidence estimation through relative comparisons.

Figure~\ref{fig:intro-fig} illustrates our method. To estimate confidence for a language model's answers to questions $q_1$, $q_2$, ..., $q_n$, we generate confidence preference data by pairing each question $q_i$  with another question $q_j$  and asking the model, \textit{``Which question are you more confident in answering correctly, $q_i$ or $q_j$?''} We repeat this $n$ times for each question to gather pairwise confidence preferences. We then convert these preferences into confidence scores, treating this as a rank aggregation problem---determining scores or rankings from a set of partial and potentially inconsistent comparisons. Leveraging well-established solutions to rank aggregation like Elo rating~\citep{elo_ratings}, Bradley-Terry~\citep{bradley_terry}, and TrueSkill~\citep{true_skill}, we translate these relative judgments of confidence into confidence scores.

We compare relative confidence estimation to state-of-the-art absolute confidence estimation methods. For absolute confidence estimates, we study direct prompting---eliciting model confidence through a single prompt---and self-consistency prompting---repeatedly prompting the model for its confidence and aggregating the results into a single score through post-processing~\citep{xiong2023can}. 

Our goal is to produce reliable confidence estimates that can allow users to detect potentially incorrect answers from the model, so we study the selective classification AUC which measures how accurate the model is if it is allowed to abstain on some (low-confidence) examples. Additionally, we also report the AUROC (Table~\ref{tab:avg_auroc_results}) to understand how well confidence scores can distinguish between correct and incorrect examples.
We evaluate relative confidence estimation on five state-of-the-art models—--Llama 3.1 405B, GPT-4, Gemini 1.5 Pro, GPT-4o, and Claude 3.5 Sonnet---on 14 challenging multiple-choice question answering tasks (GPQA, MedQA, TruthfulQA, OpenbookQA, SIQA, and 8 diverse MMLU datasets). 

Our approach matches or outperforms both direct confidence estimation and self-consistency methods for 4 out of 5 of these models (for Claude 3.5 Sonnet relative confidences slightly underperform self-consistency methods). For GPT-4o, we see 3.2\% and 1.8\% improvements respectively in AUC. For Llama 3.1 405B, we observe a 6.1\% improvement in the selective classification AUC over direct prompting and 4.9\% gain over self-consistency. Similar improvements are observed with the other models (Section~\ref{sec:results}). Our findings highlight the efficacy of relative confidences and introduce a new way of thinking about confidence estimation. 

\section{Setup}
\textbf{Task.} We follow the experimental setup described in~\citep{surrogate-models}. For a given input $x$, let $\hat y(x)$ represent the model’s output and $y(x)$ represent the gold label. $R(\hat y, y)$ is the ground truth correctness of $\hat y(x)$. Since we work with multiple choice tasks, $R(\hat{y}, y) = \mathbbm{1}\{\hat{y}(x) = y(x)\}$. 

\begin{wrapfigure}{r}
{0.5\textwidth}
    \centering
    \vspace{-0.1in}
\includegraphics[width=0.5\textwidth]
{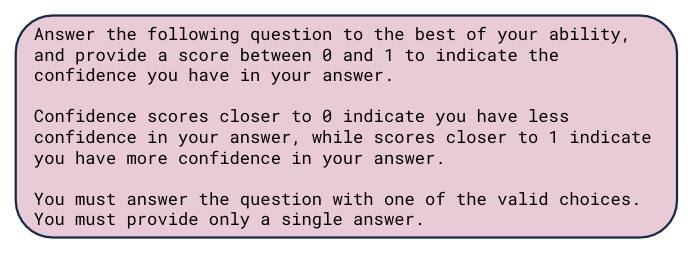}
\centering
\caption{\centering\textbf{Direct Confidence Prompt Instruction.} Asks the model to directly score its confidence in its answer to a question.}
\vspace{-0.2in}
\label{fig:direct_prompt_inst}
\end{wrapfigure}

 $C(x, \hat y) \in [0, 1]$ is the model’s confidence in $\hat y(x)$ being the correct output for $x$. Our goal is to derive reliable confidence estimates from language models---i.e. higher $C(x, \hat y)$ where $R(\hat y, y)$ is 1 and lower $C(x, \hat y)$ where $R(\hat y, y)$ is 0. Reliable confidence estimates can help prioritize high-confidence outputs and defer low-confidence cases to human experts.

\textbf{Metrics.} We measure the reliability of confidence estimates through selective classification and focus on studying the AUC~\citep{elyaniv2010foundations,liang2022helm}, area under the selective accuracy-coverage curve. The AUC measures the accuracy of a model if it is allowed to abstain on low-confidence inputs. The selective accuracy $A(c)$ is the accuracy of the model on the top $c$ fraction of examples it is most confident about. AUC is computed by aggregating the selective accuracy $A(c)$ across all $c$. We compute the AUC as described by~\citep{surrogate-models}, adding a small amount of Gaussian noise to each confidence score to allow for tie-breaking across different examples with the same confidence score. For a model with reliable confidence estimates, accuracy on a dataset should increase by abstaining on a larger fraction of low-confidence examples.

\begin{wrapfigure}{r}
{0.5\textwidth}
    \centering
    \vspace{-0.1in}
\includegraphics[width=0.5\textwidth]
{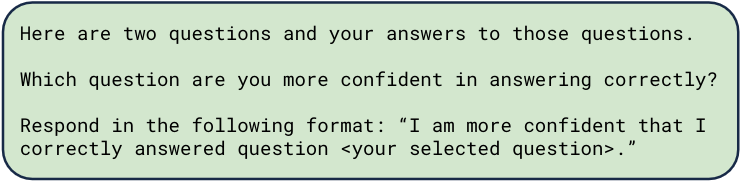}
\centering
\caption{\centering\textbf{Relative Confidence Prompt.} Asks model to compare its confidence in two questions.}
\vspace{-0.1in}
\label{fig:rel_confidence_prompt}
\end{wrapfigure}

Additionally, we also report the AUROC~\citep{hendrycks2017baseline,xiong2023can}, area under the receiver operating characteristic curve, in Appendix~\ref{appendix:auroc_results}. AUROC is a standard classification metric used to measure how well a model can separate correct and incorrect examples at different thresholds. In our setting, we use the outputted confidence scores as the thresholds for measuring AUROC. 

Expected calibration error (ECE)~\citep{Guo2017OnCO, Naeini2015ObtainingWC} is also a standard metric to measure how closely a model's confidence matches its accuracy. However, ECE does not assess a model's ability to discriminate between correct and incorrect answers---a model with accuracy 0.5 can achieve perfect ECE by outputting a confidence of 0.5 for all of its answers. Therefore, we focus our results on the AUC.

\textbf{Datasets.} We measure the quality of confidence estimates produced by the model on 14 challenging multiple-choice question answering datasets: GPQA~\citep{Rein2023GPQAAG}, MedQA~\citep{jin2021medqa}, TruthfulQA (TQA)~\citep{lin2021truthful}, CommonsenseQA (CSQA)~\citep{talmor2019commonsenseqa},
OpenbookQA (OBQA)~\citep{mihaylov2018openbook}, SIQA~\citep{siqa}, and eight diverse MMLU~\citep{hendrycks2021measuring} datasets---professional law (Law), business ethics (Ethics), conceptual physics (Physics),
econometrics (Econ), abstract algebra (Algebra), college chemistry (Chem), computer security (Security), and US Foreign Policy (Policy). We evaluate on 250 examples from the test set of each dataset. We tune the hyperparameters of our approach on a small heldout set for each task, when available, or otherwise use a fixed set of hyperparameters. See Appendix~\ref{appendix:hyperparameters} for more details.

\textbf{Models.} We evaluate our approach on five state-of-the-art models---Llama 3.1 405B~\citep{llama3.1}, GPT-4~\citep{gpt-4}, Gemini 1.5 Pro~\citep{gemini-1.5-pro}, GPT-4o~\citep{gpt-4o}, and Claude 3.5 Sonnet~\citep{claude-3.5-sonnet}.

\section{Absolute Confidence Estimation}
\label{sec:abs_conf_estimation}
Confidence estimation is often done in an \textit{absolute} setting, where a model assesses its confidence $C(x, \hat y)$ independently for each example $x$. Using a model's log probabilities as a measure of its confidence is a common absolute confidence estimation technique. We focus on \textit{linguistic} confidence estimation, where a user interacts with a model in natural language to assess its confidence, without assuming access to a model's internal representations or outputted log probabilities. Linguistic confidence estimation is becoming increasingly important as users interact with language models through chat interfaces, and several state-of-the-art models such as Claude 3.5 Sonnet and Gemini 1.5 Pro provide only API-level access to users.

We compare relative confidence estimation to two popular absolute linguistic confidence estimation methods: 

\textbf{Direct Confidence Prompting.} We zero-shot prompt the language model with an instruction to answer the question and provide a confidence estimate for that answer. Both the answer and confidence are outputted in a single generation, greedily with $T=0$. ~\cite{surrogate-models} study several different instructions for direct confidence prompting including asking the model to rate its confidence on different numerical scales, to reason about its confidence level with a chain of thought, and describe its confidence in words (e.g. ``not sure'', ``sure'', and ``very sure''). We use the direct confidence prompt from ~\cite{surrogate-models} resulting in the highest selective classification AUC across multiple language models. This prompt asks the model to rate its confidence on a scale of 0-1 and provides fake few-shot examples to allow the model to better understand the task. See Figure~\ref{fig:direct_prompt_inst} for the prompt instruction and Appendix~\ref{appendix:prompts} for the full prompt.

\textbf{Self-Consistency Confidence Prompting.} ~\cite{xiong2023can} present an extension to direct confidence prompting where motivated by work on self-consistency prompting~\citep{Wang2022SelfConsistencyIC}, multiple answers and confidences are sampled for a given question to get a more robust confidence estimate. These answers and confidences are aggregated via a post-processing procedure to produce a single answer and confidence score from the samples. See ~\cite{xiong2023can} for more details on the aggregation procedure. We follow the same procedure as ~\cite{xiong2023can}---prompting the model multiple times per question to sample different answers and confidences using the prompt in Figure~\ref{fig:direct_prompt_inst} (full prompt in Appendix~\ref{appendix:prompts}), then aggregating these samples through their post processing technique. We sample at $T=0.7$ and report results for $15$ samples.

\section{Relative Confidence Estimation}
Linguistic confidence estimation, where a model is prompted to assess its own confidence, is typically done through absolute estimation methods, in which the model independently gauges its confidence for each question. However, without clear training examples demonstrating how to estimate confidence, the model may struggle to distinguish between different confidence levels (e.g., 85\% vs. 90\%) and generate appropriate scores. In contrast, it may be easier for the model to compare its confidence across different questions, making a simpler, binary judgment about whether it is \textit{more} or \textit{less} confident in answering one question versus another. This approach provides more grounding, as confidence is evaluated relative to another question, rather than globally assessed via a direct score. By aggregating many such relative comparisons, we can still derive global confidence estimates (e.g., determining whether a question is one the model is highly confident in answering correctly).

We propose relative confidence estimation, where the model compares pairs of questions, along with its answers, and provides preference judgments on which question it is \textit{more} confident in answering correctly. Given a set of $m$ questions and their corresponding answers, our task is to elicit pairwise confidence preferences and use these preferences to derive meaningful confidence scores for each question. This process involves two stages: Confidence Preference Data Generation (Section~\ref{subsec:conf_pref_data_gen}) and Rank Aggregation (Section~\ref{subsec:rank_aggregation}).




\begin{figure*}[t]
    \centering
    \begin{minipage}{0.45\textwidth}
        \centering
        \begin{algorithm}[H]
        \caption{Confidence Preference \\ Data Generation}
        \label{alg:conf_pref_data}
        \begin{algorithmic}[1]
        \State \textbf{Input:} $Q = \{q_1, q_2, \dots, q_m\}$
        \State $\text{pref\_data} \gets \emptyset$
        \For{each $q_i \in Q$}
            \For{$k = 1$ to $n$}
                \State Randomly select $q_j \in Q \setminus q_i$
                \State winner $\gets$ Model(prompt, $q_i$, $q_j$)
                \If{winner = $q_i$} 
                    Append $(i, j)$ to $\text{pref\_data}$
                \Else $\text{ Append} (j, i)  \text{ to pref\_data}$
                \EndIf
            \EndFor
        \EndFor
        \State \textbf{Output:} $\text{pref\_data}$
        \end{algorithmic}
        \end{algorithm}
    \end{minipage}\hfill
    \begin{minipage}{0.5\textwidth}
        \centering
        \begin{algorithm}[H]
        \caption{Elo Rating}
        \label{alg:elo_scoring}
        \begin{algorithmic}[1]
        \State \textbf{Input:} $\text{pref\_data}$, $K$, $n$
        \State $S = [1000, 1000, ..., 1000]$
        \For{$i = 1$ to $\text{num\_iters}$}
            \For{each $d$ in $\text{pref\_data}$}
                \State $(w, l) \gets d$
                \State $P(w \text{ wins}) \gets \dfrac{1}{1 + 10^{(S[l] - S[w])/K}}$
                \State $P(l \text{ wins}) \gets \dfrac{1}{1 + 10^{(S[w] - S[l])/K}}$
                \State $S[w] \gets S[w] + K \times (1 - P(w \text{ wins}))\hfill$
                \State $S[l] \gets S[l] - K \times P(l \text{ wins})$
            \EndFor
        \EndFor
        \State \textbf{Output:} $S$
        \end{algorithmic}
        \end{algorithm}
    \end{minipage}
    
\end{figure*}
\subsection{Confidence Preference Data Generation}
\label{subsec:conf_pref_data_gen}
To generate confidence preference data, we employ the following procedure: for each question $i$, pair it with a randomly selected question $j\neq i$. The model compares the two questions, alongside its answers to the questions, and is then asked which one it feels more confident about answering correctly (Figure~\ref{fig:rel_confidence_prompt}). The answer for each question is obtained using the same prompt used for direct confidence prompting (Figure~\ref{fig:direct_prompt_inst}).
This process is repeated $n$ times for each question $i$, pairing it with different questions $j$ and recording the model's preferences (Algorithm~\ref{alg:conf_pref_data}).
The result is a list of confidence preference judgments $(i, j)$ indicating the model was more confident in answering question $i$ than question $j$ (or $(j, i)$ if model preferred question $j$ to $i$). 

Once this data is gathered, we move to the next step: aggregating these preferences to rank questions by confidence and using this to produce confidence scores (Section~\ref{subsec:rank_aggregation}). 

\subsection{Rank Aggregation}
\label{subsec:rank_aggregation}
Confidence preference data provides partial rankings of questions based on confidence. For instance, given questions 1, 2, and 3, the model may indicate 3 > 2 and 2 > 1. These partial rankings can be aggregated into a total ordering of questions by confidence, enabling the derivation of question-level confidence scores. This process, known as \textit{rank aggregation}, is well-studied in social choice theory for voting, consensus formation, and preference aggregation~\citep{arrow-social-choice, Tideman1987IndependenceOC, kemeny-young, Dwork2001RankAM}.

\begin{wrapfigure}{r}{0.5\textwidth} 
\vspace{-0.3in}
\begin{minipage}{0.5\textwidth}
    \centering
    \begin{algorithm}[H]
    \caption{Bradley-Terry MLE}
    \label{alg:bradley_terry}
    \begin{algorithmic}[1]
    \Function{bradley\_terry\_ll}{$\theta, \text{pref\_data}$}
        \State $S \gets exp(\theta)$
        \State $\ell(\theta; \text{pref\_data}) = 0$
        \For{each $d$ in $\text{pref\_data}$}
            \State $(w, l) \gets d$
            \State $P(w \text{ wins}) \gets \dfrac{S[w]}{S[w] + S[l]}$
            \State $\ell(\theta; \text{pref\_data}) \gets \ell(\theta; \text{pref\_data})$ 
\State \hspace{2.2em} $+ \log P(w \text{ wins})$
        \EndFor
        \State \textbf{Output:} $-\ell(\theta; \text{pref\_data}) + \frac{\lambda}{2} \sum_{i=1}^{n} \theta_i^2$
    \EndFunction
    \State
    \State \textbf{Input:} $\text{pref\_data}$
    \State $\theta = [0, 0, ..., 0]$
    \State $\text{minimize}(\text{bradley\_terry\_ll}$, $\theta$, $\text{pref\_data}$, $\text{BFGS})$
    \State $S \gets exp(\theta)$
    \State \textbf{Output:} $S$
    \end{algorithmic}
    \end{algorithm}
\end{minipage}\hfill
\vspace{-0.26in}
\end{wrapfigure} The ideal ranking would place all correctly answered questions above incorrectly answered ones, reflecting a calibrated model's confidences. With a complete set of noiseless comparisons--—where correctly answered questions are consistently preferred—--a total ordering could be derived by straightforward sorting. However, our confidence preference data is \textit{noisy} (e.g., incorrectly answered questions are sometimes preferred), \textit{inconsistent} (e.g., occasional circular preferences among questions), and \textit{incomplete} (limited to \textit{n} comparisons per question for tractability).

Given these challenges, we aim to approximate the best total ordering that represents the confidence preference data while being robust to noise, inconsistency, and incompleteness. While finding the optimal total ordering (Kemeny-optimal solutions~\citep{kemeny-young}) is NP-hard, efficient approximation algorithms can provide practical solutions.

We explore three popular algorithms to perform rank aggregation and assign confidence scores based on our preference data: Elo rating, TrueSkill, and Bradley-Terry. These algorithms are typically used to score player skill levels in tournament-style games based on matchup data. 
In this setting, each question is treated as a ``player'' engaging in matchups with other questions, where the model’s confidence preferences dictate the outcomes of these matches.

\noindent\textbf{Elo Rating.}
Elo rating~\citep{elo_ratings} is commonly used in games like chess and leverages matchup data between players to iteratively update player ratings in an online learning fashion. We start by assigning all questions identical scores. For any pair of questions $i$ and $j$, the probability of $i$ ``winning'' the matchup is modeled as a logistic function of $i$ and $j$'s current scores $s_i$, $s_j$. $K$ determines how sensitive the player scores are to match outcomes. 
\begin{equation}
    P(i \text{ wins}) = \frac{1}{1 + 10^{(s_i-s_j)/K}}
\end{equation}

After each matchup the scores are adjusted based on how significantly the estimated win probabilities deviated from the true outcome (i.e. the model’s preference)---surprising outcomes (low-confidence wins) lead to more substantial score changes. We iterate over the confidence preference data multiple times to ensure score convergence. See Algorithm~\ref{alg:elo_scoring} for more details.

\noindent \textbf{TrueSkill.} TrueSkill~\citep{true_skill} is a Bayesian model designed for ranking players in competitive games. It is an extension of the Elo rating system that represents each player's skill score as a normal distribution, with the mean ($\mu$) indicating the best estimate of their current score and the variance ($\sigma$) reflecting the model's uncertainty about
that score. After each matchup between a pair of questions, the mean and variance of each question's scores are updated based on the difference between the expected result and the true outcome. The TrueSkill model uses factor graphs to represent the probabilistic relationships between player skill levels. A belief propagation algorithm is used on the factor graph to update beliefs about players' skills based on match outcomes. As more matchup data is processed for each question, the uncertainty ($\sigma$) decreases, refining the estimate of the question’s score over time. We leverage the trueskill Python package as the implementation of this technique. \\\\
\noindent\textbf{Bradley-Terry.} The Bradley-Terry model~\citep{bradley_terry} is a probabilistic framework for modeling pairwise comparisons, commonly used in ranking tasks. It provides an alternate means of modeling the probability of question $i$ winning a matchup against question $j$, based on their underlying scores. Bradley-Terry estimates the probability that question $i$ wins over question $j$ as:\\
\begin{equation}
    P(i \text{ wins}) = \frac{s_i}{s_i + s_j}
\end{equation}
where $s_i$ and $s_j$ are the scores for question $i$ and $j$. These scores are optimized using maximum likelihood estimation (MLE), with L2 regularization applied to control for overfitting and mitigate the impact of noisy comparisons. Bradley-Terry uses a different estimate of the player win probability than Elo rating.
Additionally unlike Elo, which updates scores iteratively after each comparison, the Bradley-Terry model optimizes the scores holistically, taking all pairwise comparisons into account simultaneously. We use the BFGS algorithm to perform this optimization. See Algorithm~\ref{alg:bradley_terry} for more details.\\\\
We optimize the rank aggregation hyperparameters using a small held out set (Appendix~\ref{appendix:hyperparameters}). Finally, we normalize the confidence scores to a range of 0-1 using min-max normalization.

\section{Results}
\label{sec:results}
\begin{figure*}[!hbt]
\includegraphics[scale=0.55]{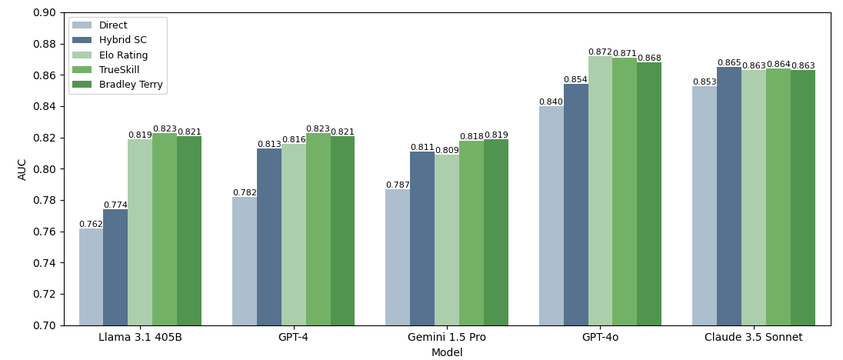}
    \caption{\textbf{Selective Classification AUC Across Models.} For each model, we plot the selective classification AUC averaged across the 14 tasks for each confidence estimation method. The absolute confidence estimation baselines---direct prompting (Direct) and self-consistency (Hybrid SC)---are indicated in blue, while relative confidence estimation with different rank aggregation methods is in green (Elo Rating, TrueSkill, Bradley-Terry). For Llama 3.1 405B, GPT-4, Gemini 1.5 Pro, and GPT-4o, relative confidence estimates outperform both the direct and hybrid SC absolute confidence baselines. For Claude 3.5 Sonnet, relative confidences outperform direct prompting but slightly underperform self-consistency prompting.
    \vspace{-0.1in}}
    \label{fig:auc_plot}
\end{figure*}
\textbf{Relative confidences outperform absolute confidences.} We compare relative confidence estimates with absolute confidence estimates using direct prompting and self-consistency prompting, and report gains over the best relative confidence approach for each model. Across 14 datasets, relative confidence estimates boost AUC over direct prompting by 6.1\% for Llama 3.1 405B, 4.1\% for GPT-4, 3.2\% for Gemini 1.5 Pro, 3.2\% for GPT-4o, and 1.1\% for Claude 3.5 Sonnet (Figure~\ref{fig:auc_plot}). Compared to self-consistency prompting, relative confidence raises AUC by 4.9\% for Llama 3.1 405B, 1.0\% for GPT-4, 0.8\% for Gemini 1.5 Pro, and 1.8\% for GPT-4o (Figure~\ref{fig:auc_plot}). For Claude 3.5 Sonnet, relative confidences slightly underperform self-consistency prompting (by 0.1\%). See Table~\ref{tab:llama_auc_results} and Table~\ref{tab:gpt4o_auc_results} respectively for the dataset-level AUC results on Llama 3.1 405B and GPT-4o, and Appendix~\ref{appendix:auc_results} for dataset-level results for the other models. Overall, relative confidence improves confidence estimation for 4 of the 5 state-of-the-art models, with Llama 3.1 405B seeing the largest gains, followed by the GPT-4 models and Gemini 1.5 Pro. While confidence estimates for Claude 3.5 Sonnet also improve over direct prompting, the gains are smaller due to Sonnet’s ability to make good absolute judgments of confidence.\\\\
\noindent \textbf{Does chain of thought improve confidence estimates?}
We experiment with augmenting relative confidence judgments in GPT-4o with chains of thought (CoTs). We update the relative confidence prompt for confidence preference data generation (Algorithm~\ref{alg:conf_pref_data}) by asking the model to reason about which question it is more confident in (Appendix~\ref{appendix:prompts}). We apply Elo rating, the best rank aggregation algorithm for GPT-4o, to the CoT confidence preference data to generate confidence scores. However, the CoT confidence estimates fail to improve performance and lead to worse outcomes when the model hallucinates evidence, becoming confident in both options. Overall, incorporating CoTs slightly decreases GPT-4o’s AUC averaged over datasets, from 87.2\% to 86.8\% while also requiring more inference-time compute.
\begin{table}[t!]
\centering
\begin{tabular}{@{}cccc|ccc@{}}
\toprule
Category & Dataset & Direct & Hybrid SC & Elo Rating & TrueSkill & Bradley-Terry \\ \midrule
 & GPQA & 0.356 & 0.293 & 0.453 & \cellcolor[HTML]{BAEFFE}\textbf{0.454} & 0.451 \\
 & MedQA & 0.864 & 0.859 & 0.914 & \cellcolor[HTML]{BAEFFE}\textbf{0.918} & 0.915 \\
 & OBQA & 0.926 & 0.959 & \cellcolor[HTML]{BAEFFE}\textbf{0.970} & 0.969 & 0.968 \\
 & Physics & 0.862 & 0.793 & 0.907 & 0.934 & \cellcolor[HTML]{BAEFFE}\textbf{0.938} \\
 & Algebra & 0.378 & 0.448 & 0.467 & 0.466 & \cellcolor[HTML]{BAEFFE}\textbf{0.476} \\
 & Chem & 0.585 & 0.486 & 0.747 & \cellcolor[HTML]{BAEFFE}\textbf{0.751} & 0.746 \\
\multirow{-7}{*}{STEM} & Security & 0.861 & 0.899 & 0.895 & \cellcolor[HTML]{BAEFFE}\textbf{0.910} & 0.908 \\ \midrule
 & Law & 0.749 & 0.747 & 0.813 & \cellcolor[HTML]{BAEFFE}\textbf{0.834} & 0.825 \\
 & Ethics & 0.889 & \cellcolor[HTML]{BAEFFE}\textbf{0.963} & 0.922 & 0.922 & 0.917 \\
 & Econ & 0.711 & 0.703 & \cellcolor[HTML]{BAEFFE}\textbf{0.778} & 0.770 & 0.748 \\
\multirow{-4}{*}{Social Sciences} & Policy & 0.961 & \cellcolor[HTML]{BAEFFE}\textbf{0.993} & 0.989 & 0.987 & 0.990 \\ \midrule
 & TQA & 0.861 & \cellcolor[HTML]{BAEFFE}\textbf{0.899} & 0.876 & 0.874 & 0.877 \\
 & CSQA & 0.837 & \cellcolor[HTML]{BAEFFE}\textbf{0.920} & 0.865 & 0.868 & 0.870 \\
\multirow{-3}{*}{\begin{tabular}[c]{@{}c@{}}Commonsense\\ Reasoning\end{tabular}} & SIQA & 0.830 & \cellcolor[HTML]{BAEFFE}\textbf{0.879} & 0.868 & 0.867 & 0.871 \\ \midrule
 & Average & 0.762 & 0.774 & 0.819 & \cellcolor[HTML]{BAEFFE}\textbf{0.823} & 0.821 \\ \bottomrule
\end{tabular}
\caption{\textbf{Llama 3.1 405B AUCs All Methods.} We show the dataset-level results for Llama 3.1 405B, for the Direct and Hybrid SC absolute confidence baselines and for relative confidence estimation with different rank aggregation methods (Elo Rating, TrueSkill, Bradley-Terry). Relative confidences outperform absolute confidences for all STEM datasets, whereas absolute confidences with self-consistency (Hybrid SC) work best for commonsense reasoning tasks. Overall, relative confidences with TrueSkill rank aggregation lead to a 6.1\% improvement over direct prompting and a 4.9\% improvement over self-consistency prompting.}
\label{tab:llama_auc_results}
\end{table}\\\\
\noindent \textbf{How important are answers in determining confidence?} 
We investigate how important it is for a model to see its own answer to a question in order to gauge its confidence level in correctly answering the question. To assess this, we modify the relative confidence prompt, asking GPT-4o to judge which of the two questions is more \textit{difficult} for it to answer correctly, without providing it access to its own answers to these questions. See Appendix~\ref{appendix:prompts} for the exact prompt. We then apply the same rank aggregation methods to this difficulty preference data and produce confidence scores. This approach drops the average AUC for relative confidence estimation with Elo rating by 5.3\% from 87.2\% to 81.9\%, emphasizing that access to its own answers significantly enhances the model’s relative confidence judgments. Nevertheless, even without answers, relative confidence judgments are only 2.1\% less reliable than absolute confidence assessments with answers (81.9\% vs 84\%), suggesting that models are still reasonably good at judging a question's difficulty, even before answering it. \\\\

\begin{wraptable}{r}{0.55\textwidth}
\begin{tabular}{@{}ccc@{}}
\toprule
\# Model Calls & $\%$ Gains GPT-4o & $\%$ Gains Llama 3.1 \\ \midrule
5 & $0.9\%$ & $2.2\%$ \\
10 & $1.8\%$ & $3.2\%$ \\
15 & $1.8\%$ & $4.9\%$ \\ \bottomrule
\end{tabular}
\caption{\textbf{Gains by scaling up comparisons.} We report the gains of relative confidence estimation over self-consistency across different numbers of model calls. }
\label{tab:scaling_comparisons}
\end{wraptable}

\noindent \textbf{Does scaling up comparisons help?} We hypothesize that increasing the number of relative confidence comparisons per question would lead to a better ranking of questions by confidence, and more reliable confidence scores. To test this, we scale up the number of judgments, going from 5 to 10 to 15 model calls per question. To ensure a fair comparison based on compute, we use a self-consistency baseline with the same number of model calls per confidence estimate (Section~\ref{sec:abs_conf_estimation}). We report improvements based on the best rank aggregation method for each model in Table~\ref{tab:scaling_comparisons}.
Even for a small number of model calls, relative confidences show improvements over self-consistency prompting. Further scaling up the number of relative confidence comparisons per question increases the improvements of relative confidence estimation over self-consistency prompting. However, as seen with GPT-4o, for some models further scaling model calls may show diminishing returns due to inherent noise in the model’s confidence preferences.
\begin{table}[]
\centering
\begin{tabular}{@{}cccc|ccc@{}}
\toprule
Category & Dataset & Direct & Hybrid SC & Elo Rating & TrueSkill & Bradley-Terry \\ \midrule
 & GPQA & 0.480 & 0.421 & \cellcolor[HTML]{BAEFFE}\textbf{0.530} & 0.528 & 0.522 \\
 & MedQA & 0.923 & 0.931 & \cellcolor[HTML]{BAEFFE}\textbf{0.944} & 0.943 & 0.943 \\
 & OBQA & 0.971 & 0.983 & \cellcolor[HTML]{BAEFFE}\textbf{0.987} & 0.986 & \cellcolor[HTML]{BAEFFE}\textbf{0.987} \\
 & Physics & 0.898 & 0.914 & 0.940 & 0.944 & \cellcolor[HTML]{BAEFFE}\textbf{0.946} \\
 & Algebra & 0.655 & \cellcolor[HTML]{BAEFFE}\textbf{0.743} & 0.722 & 0.710 & 0.694 \\
 & Chem & 0.741 & 0.700 & 0.795 & \cellcolor[HTML]{BAEFFE}\textbf{0.806} & 0.802 \\
\multirow{-7}{*}{STEM} & Security & 0.880 & 0.913 & \cellcolor[HTML]{BAEFFE}\textbf{0.930} & 0.927 & 0.922 \\ \midrule
 & Law & 0.859 & \cellcolor[HTML]{BAEFFE}\textbf{0.872} & \cellcolor[HTML]{BAEFFE}\textbf{0.872} & 0.867 & 0.867 \\
 & Ethics & 0.960 & \cellcolor[HTML]{BAEFFE}\textbf{0.969} & 0.962 & 0.962 & 0.959 \\
 & Econ & 0.799 & 0.824 & \cellcolor[HTML]{BAEFFE}\textbf{0.837} & 0.833 & 0.833 \\
\multirow{-4}{*}{Social Sciences} & Policy & 0.962 & 0.965 & \cellcolor[HTML]{BAEFFE}\textbf{0.983} & \cellcolor[HTML]{BAEFFE}\textbf{0.983} & 0.980 \\ \midrule
 & TQA & 0.906 & \cellcolor[HTML]{BAEFFE}\textbf{0.935} & 0.908 & 0.911 & 0.911 \\
 & CSQA & 0.864 & \cellcolor[HTML]{BAEFFE}\textbf{0.900} & 0.886 & 0.887 & 0.884 \\
\multirow{-3}{*}{\begin{tabular}[c]{@{}c@{}}Commonsense\\ Reasoning\end{tabular}} & SIQA & 0.855 & 0.884 & 0.905 & 0.905 & \cellcolor[HTML]{BAEFFE}\textbf{0.908} \\ \midrule
 & Average & 0.840 & 0.854 & \cellcolor[HTML]{BAEFFE}\textbf{0.872} & 0.871 & 0.868 \\ \bottomrule
\end{tabular}
\caption{\textbf{GPT-4o AUCs All Methods.} We show the dataset-level results for GPT-4o, for the Direct and Hybrid SC absolute confidence baselines and for relative confidence estimation with different rank aggregation methods (Elo Rating, TrueSkill, Bradley-Terry). Relative confidences outperform absolute confidences for the majority of STEM and social science datasets, while absolute confidences with self-consistency tend to work better for commonsense reasoning tasks.}
\label{tab:gpt4o_auc_results}
\end{table}\\\\
\noindent \textbf{Different methods for rank aggregation.} We evaluate multiple rank aggregation methods for converting relative confidence preferences into scalar scores. Relative confidence estimation with any rank aggregation method outperforms direct and self-consistency prompting (Figure~\ref{fig:auc_plot}) (except for slightly underperforming self-consistency prompting with Claude 3.5 Sonnet). While differences in the performance of the rank aggregation methods is small, TrueSkill is the best method for most models, except for Gemini 1.5 Pro where Bradley-Terry performs best and GPT-4o where Elo rating performs best. 

TrueSkill explicitly models player skill levels as probability distributions instead of single point estimates, as in Elo rating and Bradley-Terry. This allows it to capture uncertainty in each player’s skill rating and update it as they participate in more games, which may allow this method to be more robust to the noise in the relative comparison data. In general, for relative confidence estimation with a new model, we would recommend starting with TrueSkill rank aggregation. The online learning paradigm of Elo rating and TrueSkill may also be particularly suited to environments where confidence judgments accumulate over time, leading to more refined confidence estimates (i.e. confidences of a medical chatbot improving as it helps more patients), in contrast to Bradley-Terry where confidence scores are optimized over the full dataset of judgments at once.

\section{Related Work}
\textbf{Confidence Estimation.}  Recent studies have explored confidence estimation in language models. ~\citet{kadavath2022language} measure the calibration of outputted log probabilities from language models and find that models generally demonstrate good calibration on true/false and multiple-choice tasks. They also show that models can better estimate their confidence in an answer by comparing multiple answers for a given question. Our approach instead asks models to compare their confidence across different questions and finds this leads to reliable confidence estimates. ~\cite{surrogate-models} show that absolute linguistic confidence estimation (e.g. “Score your confidence from 0-1”) is a hard problem for closed models, and confidences for closed models can instead be estimated by transferring log probabilities from open models. Our work instead focuses on linguistic confidence estimates, without needing access to a model’s log probabilities. Other works on linguistic confidence estimation use self-consistency-like methods to sample multiple answers and corresponding confidences from models and aggregate them~\citep{xiong2023can}. We compare relative confidence estimation with the best performing self-consistency technique from~\cite{xiong2023can} and find that relative confidences tend to outperform self-consistency based estimates. Other approaches fine-tune language models to improve confidence estimation~\citep{lin2022teaching}, while our method elicits better estimates without requiring further training. 

\textbf{LMs as Evaluators. } Several works also use language models to evaluate the quality of a model’s responses. GPTScore~\citep{Fu2023GPTScoreEA} and LLM-as-a-judge~\citep{Zheng2023JudgingLW} use LMs to provide automated scoring or feedback on different aspects of text quality as an alternative to traditional text evaluation metrics such as ROUGE and BLEU. These approaches are similar to absolute linguistic confidence estimation (“Score your confidence from 0-1”). Other works use LMs to evaluate their responses through either a numerical score or natural language feedback to improve their own generations. This can occur through search at decoding time~\citep{Yao2023TreeOT}, prompting the model to self-correct its responses using its feedback~\citep{Madaan2023SelfRefineIR, Bai2022ConstitutionalAH}, or by aligning a model using its own reward signals~\citep{Yuan2024SelfRewardingLM}. Linguistic confidence estimation relates to self-evaluation with LMs, since we ask models to evaluate their own confidence levels. 

\textbf{Learning from Human Preference Data.} Several approaches have improved language models across diverse attributes (safety, fluency, etc.) by deriving a reward signal from human preferences. These preferences are typically framed as relative judgments by asking annotators to select their preferred output from a pair or set of responses for a given input, instead of asking them to directly score the quality of a single response~\citep{instruct-gpt, Ziegler2019FineTuningLM, Christiano2017DeepRL}. Motivated by this framing, we elicit relative confidence judgments from LMs and use these to produce more reliable confidence scores.

\textbf{Rank Aggregation.} There is a rich body of work studying the problem of rank aggregation--–converting partial orderings over a set into a better total ordering~\citep{arrow-social-choice, Tideman1987IndependenceOC, kemeny-young, Dwork2001RankAM}. This problem is common in domains such as sports and competitive games, election voting, and product recommendations. Our work leverages popular rank aggregation algorithms such as Elo rating~\citep{elo_ratings}, TrueSkill~\citep{true_skill}, and Bradley-Terry~\citep{bradley_terry} to convert the pairwise confidence preferences from a model into a total ordering of questions and corresponding answers by confidence. Other approaches such as Rank Centrality~\citep{Negahban2012RankCR} model rank aggregation through a Markov Chain and use the stationary distribution to determine the rank of each item.

\textbf{Calibration and Selective Classification.} The quality of confidence estimates is often measured through calibration---by determining how grounded the confidences are in true correctness~\citep{murphy1977reliability,degroot1983forecasters,naeini2014binary,guo2017calibration}, typically through the expected calibration error (ECE). However, the ECE cannot capture how well confidences distinguish between correct and incorrect examples: outputting the same confidence for all examples can lead to perfect ECE if the confidence matches the average model accuracy. This leads us to focus on selective classification~\citep{elyaniv2010foundations,khani2016unanimity,feng2019selective,jones2021selective} which measures if the model “knows what it doesn’t know” and can achieve high accuracy by abstaining on examples where it is uncertain.

\section{Discussion}

As users increasingly interact with language models through chat interfaces, estimating linguistic confidences by asking the model about its confidence in natural language has become increasingly important. Most current approaches rely on absolute confidence estimates, where the model is asked to judge its confidence for a question in isolation, e.g., “rate your confidence on a scale of 0-1.” However, prior work shows that models struggle with absolute confidence estimation, as they are not specifically trained to produce such estimates~\citep{zhou2023navigating}. As a result, they tend to default to a narrow range of coarse-grained confidences for most questions (e.g., 0.9, 0.95), which fail to convey meaningful distinctions in certainty to users~\citep{surrogate-models}.

In contrast, relative preferences are ubiquitous in real life, from ranking players in games to conducting A/B testing for products. Relative preferences are also highly effective in machine learning. For example, relative annotations of generation quality lead to better reward estimates in RLHF, and models are shown to be better calibrated on multiple-choice questions~\citep{kadavath2022language}, which also involve relative judgments.

Given the challenges with absolute confidence estimation, we propose a shift towards relative confidence estimation. Rather than asking models to directly generate \textit{confidence scores}, we ask them to instead provide \textit{confidence preferences} by comparing their confidence levels across pairs of questions. These preferences can then be converted into confidence scores using rank aggregation methods, such as Elo rating~\citep{elo_ratings} and the Bradley-Terry model~\citep{bradley_terry}. By framing confidence estimation as a simpler binary decision—``more confident'' or ``less confident''—we reduce the complexity of the task and eliminate the need for models to generate fine-grained confidence scores in isolation. To our best knowledge, we are the first work to approach confidence estimation through the lens of relative comparisons.

Our method is further motivated by the notion that, for any given task, questions can be ranked along a spectrum of difficulty for a given model. Harder questions, which the model is more likely to answer incorrectly, should correspond to lower confidence scores. Relative confidence estimation leverages this principle, using pairwise confidence comparisons and rank aggregation to approximate a ranking of questions by ``difficulty'', thereby producing more meaningful confidence estimates.

We show the effectiveness of relative confidence estimation over absolute confidence estimation across a broad range of question answering tasks, demonstrating improved confidence estimates for five state-of-the-art language models.

\section{Future Work}
\noindent \textbf{Eliciting Confidence Preference Data.} There can be several different ways of eliciting relative confidence judgments. Prompts could allow for ties in confidence or compare confidence across more than two questions. Kahneman-Tversky Optimization (KTO)~\citep{Ethayarajh2024KTOMA} for LM alignment 
achieves DPO~\citep{Rafailov2023DirectPO} levels of performance by using binary signals of desirability for generations. We can apply KTO to confidence preference data generation by asking for binary signals—--confident or not—--and then converting these into relative judgments, ranking “not confident” answers below “confident” ones.\\\\
\noindent \textbf{Rank Aggregation.} In this work, we explore the most popular rank aggregation methods like Elo rating~\citep{elo_ratings}, Bradley-Terry~\citep{bradley_terry}, and TrueSkill~\citep{true_skill}. Another approach to rank aggregation is to represent preference data as a graph, with nodes as questions and directed edges reflecting match outcomes between questions. Since the outcome of some of these matchups can be inconsistent and non-transitive, algorithms like Rank Centrality~\citep{Negahban2012RankCR}, PageRank~\citep{Page1999ThePC}, and Minimum Feedback Arc Set~\citep{Vahidi2024MinimumWF} could be used to reduce cycles in the graph and better manage these inconsistencies.\\\\
\noindent \textbf{Confidence Estimation for Longform Generations.} While we benchmark on multiple-choice tasks, relative confidence estimation can also extend to longform generation. Log probabilities on answer tokens are commonly used for confidence estimation in multiple-choice tasks, but token-level uncertainty doesn't translate well to longform sequences. Moreover, there may be different levels of uncertainty associated with different aspects of a longform generation, e.g. how complete a generation, vs how factual it is, etc. Relative confidence estimation could produce fine-grained confidence scores for different attributes of a longform response by adjusting the prompt for confidence preferences accordingly.\\\\
\noindent \textbf{Alignment with Relative Confidence.} Works like~\cite{Tian2023FinetuningLM} explore using absolute confidence scores to align language models for different attributes such as factuality, without human annotations (RLAIF). Since relative confidences are more calibrated than absolute confidences, we can instead use relative confidences to construct preference pairs for aligning models on different attributes. \\\\
\noindent \textbf{Curriculum Learning with Difficulty Estimates.} We also explore generating relative confidence judgments without revealing model answers (Section~\ref{sec:results}). These scores correspond to difficulty ratings, which could inform curriculum learning by first training on lower-difficulty examples.
\section{Acknowledgments}
We sincerely thank Tushar Khot for his insightful discussions and guidance on this work during our time collaborating with the Allen Institute of AI (AI2). His feedback was invaluable in shaping the early aspects of this work.
\newpage
\bibliography{conference, anthology, custom}
\bibliographystyle{tmlr}

\newpage
\appendix
\section{Appendix}
\subsection{Full AUC Results}
\label{appendix:auc_results}
\begin{table}[H]
\centering
\begin{tabular}{@{}cccc|ccc@{}}
\toprule
Category & Dataset & Direct & Hybrid SC & Elo Rating & TrueSkill & Bradley-Terry \\ \midrule
 & GPQA & 0.441 & 0.457 & 0.454 & 0.460 & \cellcolor[HTML]{BAEFFE}\textbf{0.466} \\
 & MedQA & 0.904 & \cellcolor[HTML]{BAEFFE}\textbf{0.920} & 0.893 & 0.900 & 0.901 \\
 & OBQA & 0.979 & 0.984 & 0.984 & 0.984 & \cellcolor[HTML]{BAEFFE}\textbf{0.985} \\
 & Physics & 0.909 & 0.911 & 0.925 & \cellcolor[HTML]{BAEFFE}\textbf{0.928} & 0.927 \\
 & Algebra & 0.802 & \cellcolor[HTML]{BAEFFE}\textbf{0.811} & 0.806 & 0.805 & 0.804 \\
 & Chem & 0.832 & 0.840 & \cellcolor[HTML]{BAEFFE}\textbf{0.864} & 0.854 & 0.850 \\
\multirow{-7}{*}{STEM} & Security & 0.920 & 0.930 & \cellcolor[HTML]{BAEFFE}\textbf{0.934} & 0.930 & 0.917 \\ \midrule
 & Law & 0.789 & 0.809 & 0.799 & 0.815 & \cellcolor[HTML]{BAEFFE}\textbf{0.816} \\
 & Ethics & 0.956 & 0.964 & \cellcolor[HTML]{BAEFFE}\textbf{0.971} & 0.970 & 0.966 \\
 & Econ & 0.798 & 0.822 & 0.825 & \cellcolor[HTML]{BAEFFE}\textbf{0.829} & 0.819 \\
\multirow{-4}{*}{Social Sciences} & Policy & 0.991 & \cellcolor[HTML]{BAEFFE}\textbf{0.995} & 0.982 & 0.980 & 0.982 \\ \midrule
 & TQA & 0.880 & 0.889 & \cellcolor[HTML]{BAEFFE}\textbf{0.918} & 0.917 & 0.917 \\
 & CSQA & 0.885 & \cellcolor[HTML]{BAEFFE}\textbf{0.887} & 0.873 & 0.869 & 0.872 \\
\multirow{-3}{*}{\begin{tabular}[c]{@{}c@{}}Commonsense\\ Reasoning\end{tabular}} & SIQA & 0.861 & \cellcolor[HTML]{BAEFFE}\textbf{0.897} & 0.861 & 0.856 & 0.863 \\ \midrule
 & Average & 0.853 & \cellcolor[HTML]{BAEFFE}\textbf{0.865} & 0.863 & 0.864 & 0.863 \\ \bottomrule
\end{tabular}
\caption{\textbf{Claude 3.5 Sonnet AUCs All Methods.} We show the dataset-level results for Claude 3.5 Sonnet, for the Direct and Hybrid SC absolute confidence baselines and for relative confidence estimation with different rank aggregation methods (Elo Rating, TrueSkill, Bradley-Terry). Relative confidences outperform absolute confidence baselines for 9 out of 14 datasets across STEM, social science, and commonsense reasoning. On average, relative confidences closely match the performance of the best absolute confidence methods (only 0.1\% lower AUC than self-consistency prompting).}
\label{tab:claude_auc_results}
\end{table}
\begin{table}[H]
\centering
\begin{tabular}{@{}cccc|ccc@{}}
\toprule
Category & Dataset & Direct & Hybrid SC & Elo Rating & TrueSkill & Bradley-Terry \\ \midrule
 & GPQA & 0.395 & \cellcolor[HTML]{BAEFFE}\textbf{0.424} & 0.410 & 0.409 & 0.413 \\
 & MedQA & 0.794 & \cellcolor[HTML]{BAEFFE}\textbf{0.831} & 0.725 & 0.786 & 0.792 \\
 & OBQA & 0.955 & 0.959 & 0.979 & \cellcolor[HTML]{BAEFFE}\textbf{0.985} & \cellcolor[HTML]{BAEFFE}\textbf{0.985} \\
 & Physics & 0.878 & 0.922 & 0.927 & \cellcolor[HTML]{BAEFFE}\textbf{0.936} & 0.934 \\
 & Algebra & 0.603 & 0.612 & 0.629 & \cellcolor[HTML]{BAEFFE}\textbf{0.651} & 0.640 \\
 & Chem & 0.717 & 0.762 & 0.806 & \cellcolor[HTML]{BAEFFE}\textbf{0.851} & 0.842 \\
\multirow{-7}{*}{STEM} & Security & \cellcolor[HTML]{BAEFFE}\textbf{0.868} & 0.863 & 0.850 & 0.824 & 0.837 \\ \midrule
 & Law & 0.695 & 0.727 & 0.766 & 0.776 & \cellcolor[HTML]{BAEFFE}\textbf{0.778} \\
 & Ethics & 0.903 & 0.910 & 0.949 & 0.954 & \cellcolor[HTML]{BAEFFE}\textbf{0.957} \\
 & Econ & 0.684 & 0.734 & \cellcolor[HTML]{BAEFFE}\textbf{0.747} & 0.732 & 0.736 \\
\multirow{-4}{*}{Social Sciences} & Policy & 0.961 & 0.957 & \cellcolor[HTML]{BAEFFE}\textbf{0.980} & 0.979 & 0.978 \\ \midrule
 & TQA & 0.876 & \cellcolor[HTML]{BAEFFE}\textbf{0.891} & 0.861 & 0.853 & 0.854 \\
 & CSQA & 0.835 & \cellcolor[HTML]{BAEFFE}\textbf{0.889} & 0.860 & 0.869 & 0.872 \\
\multirow{-3}{*}{\begin{tabular}[c]{@{}c@{}}Commonsense\\ Reasoning\end{tabular}} & SIQA & 0.854 & \cellcolor[HTML]{BAEFFE}\textbf{0.874} & 0.840 & 0.854 & 0.848 \\ \midrule
 & Average & 0.787 & 0.811 & 0.809 & 0.818 & \cellcolor[HTML]{BAEFFE}\textbf{0.819} \\ \bottomrule
\end{tabular}
\caption{\textbf{Gemini 1.5 Pro AUCs All Methods.} We show the dataset-level AUC results for Gemini 1.5 Pro. On average, relative confidence estimation with Bradley-Terry leads to the best AUC with a 3.2\% improvement over direct prompting and a 0.8\% improvement over self-consistency prompting.
}
\label{tab:gemini_auc_results}
\end{table}
\begin{table}[]
\centering
\begin{tabular}{@{}cccc|ccc@{}}
\toprule
Category & Dataset & Direct & Hybrid SC & Elo Rating & TrueSkill & Bradley-Terry \\ \midrule
 & GPQA & 0.393 & 0.383 & 0.377 & \cellcolor[HTML]{BAEFFE}\textbf{0.404} & 0.394 \\
 & MedQA & 0.841 & \cellcolor[HTML]{BAEFFE}\textbf{0.893} & 0.870 & 0.875 & 0.864 \\
 & OBQA & 0.966 & 0.979 & \cellcolor[HTML]{BAEFFE}\textbf{0.990} & \cellcolor[HTML]{BAEFFE}\textbf{0.990} & 0.989 \\
 & Physics & 0.818 & 0.851 & 0.908 & \cellcolor[HTML]{BAEFFE}\textbf{0.918} & 0.917 \\
 & Algebra & 0.587 & 0.650 & 0.642 & 0.651 & \cellcolor[HTML]{BAEFFE}\textbf{0.663} \\
 & Chem & 0.682 & 0.774 & 0.797 & \cellcolor[HTML]{BAEFFE}\textbf{0.805} & 0.795 \\
\multirow{-7}{*}{STEM} & Security & 0.911 & 0.916 & \cellcolor[HTML]{BAEFFE}\textbf{0.933} & 0.927 & 0.922 \\ \midrule
 & Law & 0.716 & 0.741 & 0.722 & 0.753 & \cellcolor[HTML]{BAEFFE}\textbf{0.754} \\
 & Ethics & 0.870 & \cellcolor[HTML]{BAEFFE}\textbf{0.915} & 0.908 & 0.914 & 0.911 \\
 & Econ & 0.634 & 0.638 & 0.717 & 0.714 & \cellcolor[HTML]{BAEFFE}\textbf{0.725} \\
\multirow{-4}{*}{Social Sciences} & Policy & 0.959 & \cellcolor[HTML]{BAEFFE}\textbf{0.973} & 0.970 & 0.971 & 0.971 \\ \midrule
 & TQA & 0.892 & \cellcolor[HTML]{BAEFFE}\textbf{0.926} & 0.865 & 0.869 & 0.872 \\
 & CSQA & 0.831 & \cellcolor[HTML]{BAEFFE}\textbf{0.868} & 0.835 & 0.841 & 0.837 \\
\multirow{-3}{*}{\begin{tabular}[c]{@{}c@{}}Commonsense\\ Reasoning\end{tabular}} & SIQA & 0.851 & 0.872 & 0.886 & \cellcolor[HTML]{BAEFFE}\textbf{0.888} & 0.887 \\ \midrule
 & Average & 0.782 & 0.813 & 0.816 & \cellcolor[HTML]{BAEFFE}\textbf{0.823} & 0.821 \\ \bottomrule
\end{tabular}
\caption{\textbf{GPT-4 AUCs All Methods.} For GPT-4, relative confidences with TrueSkill lead to the best average AUC with a 4.1\% improvement over direct prompting and a 1.0\% improvement over self-consistency.}
\label{tab:gpt4_auc_results}
\end{table}

\subsection{Average AUROC Results}
\label{appendix:auroc_results}
\begin{table}[H]
\centering
\begin{tabular}{@{}ccc|ccc@{}}
\toprule
Model & Direct & Hybrid SC & Elo Rating & TrueSkill & Bradley-Terry \\ \midrule
Llama 3.1 405B & 0.575 & 0.774 & 0.849 & \cellcolor[HTML]{BAEFFE}\textbf{0.856} & 0.852 \\
GPT-4 & 0.642 & \cellcolor[HTML]{BAEFFE}\textbf{0.730} & 0.708 & 0.719 & 0.713 \\
Gemini 1.5 Pro & 0.627 & 0.700 & 0.689 & \cellcolor[HTML]{BAEFFE}\textbf{0.713} & 0.712 \\
GPT-4o & 0.698 & \cellcolor[HTML]{BAEFFE}\textbf{0.774} & 0.762 & 0.763 & 0.758 \\
Claude 3.5 Sonnet & 0.685 & \cellcolor[HTML]{BAEFFE}\textbf{0.726} & 0.711 & 0.713 & 0.713 \\ \midrule
Average Across Models & 0.645 & 0.741 & 0.744 & \cellcolor[HTML]{BAEFFE}\textbf{0.753} & 0.749 \\ \bottomrule
\end{tabular}
\caption{\textbf{Model AUROCs.} Relative confidences with TrueSkill lead to the best average AUROC for 2 out of 5 models, and a 10.8\% gain over direct prompting and a 1.2\% gain over self-consistency across all models.}
\label{tab:avg_auroc_results}
\end{table}

\subsection{Hyperparameters}
\label{appendix:hyperparameters}
Following are the hyperparameters involved for each rank aggregation method of relative confidence estimation.

\textbf{Elo rating.} initial scores, $K$, $\#$ iterations

\textbf{TrueSkill.} $\mu$, $\sigma$, $\beta$, $\tau$

\textbf{Bradley-Terry.} maximum $\#$ iterations, $\lambda$ for regularization

We use the following fixed set of hyperparameters for datasets which do not have a sufficient validation set for hyperparameter tuning of a hundred examples or more beyond their test set.

\begin{table}[H]
\centering
\begin{tabular}{@{}ccc|cccc|cc@{}}
\toprule
\multicolumn{3}{c|}{Elo Rating} & \multicolumn{4}{c|}{TrueSkill} & \multicolumn{2}{c}{Bradley-Terry} \\ \midrule
Initial Score & \textit{K} & \# iterations & $\mu$ & $\sigma$ & $\beta$ & $\tau$ & max \# iterations & $\lambda$ \\ \midrule
1000 & 400 & 1 & 25.0 & $\frac{\mu}{3.0}$ & $\frac{\mu}{6.0}$ & $\frac{\mu}{300.0}$ & 5 & 0.01 \\ \bottomrule
\end{tabular}
\caption{\textbf{Rank Aggregation Hyperparameter Values.}}
\label{tab:rank_agg_hyperparams}
\end{table}

For the datasets which have a hundred or more examples in their train or validation sets, we select a hundred examples to use for tuning the following hyperparameters to achieve the best AUC on this held-out set. 

\begin{table}[H]
\centering
\begin{tabular}{@{}c|cc@{}}
\toprule
Algorithm & Parameter & Values \\ \midrule
Elo Rating & \# iters & {[}1-20{]} \\ \midrule
\multirow{3}{*}{TrueSkill} & $\sigma$ & {[}$\frac{\mu}{3.0}$, $\frac{\mu}{2.5}$, $\frac{\mu}{2.2}$, $\frac{\mu}{2.0}${]} \\
 & $\beta$ & {[}$\frac{\mu}{6.0}$, $\frac{\mu}{5.0}$, $\frac{\mu}{4.0}$, $\frac{\mu}{3.0}${]} \\
 & $\tau$ & {[}$\frac{\mu}{300.0}$, $\frac{\mu}{250.0}$, $\frac{\mu}{200.0}$, $\frac{\mu}{150.0}${]} \\ \midrule
Bradley-Terry & max \# iters & {[}1-20{]} \\ \bottomrule
\end{tabular}
\caption{\textbf{Rank Aggregation Hyperparameter Ranges.}}
\end{table}

\subsection{Prompts}
\label{appendix:prompts}
\begin{center}
{\textbf{Linguistic Confidence Prompt}}
  \fbox{\parbox{0.8\textwidth}{%
    Answer the following question to the best of your ability, and provide a score between 0 and 1 to indicate the confidence you have in your answer. Confidence scores closer to 0 indicate you have less confidence in your answer, while scores closer to 1 indicate you have more confidence in your answer. You must answer the question with one of the valid choices. You must provide only a single answer. \\\\
    Question: This is a question\\
    (A) first answer\\
    (B) second answer\\
    (C) third answer\\
    (D) fourth answer\\
    (E) fifth answer\\
    Answer: (D)\\
    Confidence: 0.4\\\\
    Question: This is another question\\
    (A) first answer\\
    (B) second answer\\
    (C) third answer\\
    (D) fourth answer\\
    (E) fifth answer\\
    Answer: (A)\\
    Confidence: 0.7
  }}
\end{center}

\begin{center}
{\textbf{CoT Relative Confidence Prompt}}
  \fbox{\parbox{0.8\textwidth}{%
Here are two questions and your answers to those questions. Which question are you more confident in answering correctly and why? Respond in the following format: `I am more confident that I correctly answered question <your selected question>, because <your reasoning>.'   
  }}
\end{center}

\begin{center}
{\textbf{Difficulty Prompt}}
  \fbox{\parbox{0.8\textwidth}{%
Here are two questions. Which question is more difficult? Respond in the following format: `<your selected question> is more difficult.' 
  }}
\end{center}

\end{document}